\documentclass[a4, 10 pt, conference]{ieeeconf}
\usepackage{url,graphicx,subfigure,float,pxfonts}

\title{\LARGE \bf Development of a Cost-efficient Autonomous MAV\\
for an Unstructured Indoor Environment}

\author{ $^1$Serge Kernbach, $^2$Kristof Jebens \\
$^{1,2}$ Institute \footnote{sds} Parallel and Distributed Systems, University of Stuttgart, \\Universit\"atstr. 38, 70569 Stuttgart, Germany \\
{\small $^1$serge.kernbach@ipvs.uni-stuttgart.de}, {\small $^2$former member of the institute}}

\begin{document}

\maketitle
\thispagestyle{empty}
\pagestyle{empty}

\begin{abstract}
Performing rescuing and surveillance operations with autonomous ground and aerial vehicles become more and more apparent task. Involving unmanned robot systems allows making these operations more efficient, safe and reliable especially in hazardous areas. This work is devoted to the development of a cost-efficient micro aerial vehicle in a quadrocopter shape for developmental purposes within indoor scenarios. It has been constructed with off-the-shelf components available for mini helicopters. Additional sensors and electronics are incorporated into this aerial vehicle to stabilize its flight behavior and to provide a capability of an autonomous navigation in a partially unstructured indoor environment.
\end{abstract}

\section{Introduction}

Collective robotics is a young research field, which is devoted to different ground, aerial and underwater systems~\cite{Kernbach11-HCR}. Advantages of collective robotic are a high reliability, extended spatial properties, collective efficiency, which define application fields such as search and rescue operations, environmental monitoring or surveillance scenarios~\cite{Howden2009}. Especial attention is paid to micro aerial vehicles (MAV) due to their small size, low cost and a potential of creating a large-scale system~\cite{Zufferey10}.

The goal of this work is to develop an autonomous cost-efficient MAV, which is capable of indoor flight in a partially unstructured environment. MAVs have several advantages over ground vehicles due to their capability of passing objects in the third dimension rather than finding a suitable 2D route around an obstacle. This feature considerably improves the use of autonomous vehicles and also decreases computational complexity and difficulty for maneuvering to a point of interest. 

Since a quadrocopter is like a Sikorsky helicopter\footnote{Named after Igor Sikorsky; developer of helicopter with single main and tail rotor} a very instable aerial vehicle, it needs several stabilization mechanisms, which always keep the vehicle in the upright position~\cite{McKerrow04}. Without a kind of stabilization the MAV flights quickly towards one direction and can flip in the air. A very simple stabilization can be accomplished using two gyroscope sensors one for the nick and one for the roll axis with two independent PID\footnote{Proportional-integral-derivative controller (PID controller)} controllers to minimize the angular velocity~\cite{Tayebi06}. Although this kind of regulation provides relatively stable flight there are many complex enhancements possible.

Since this work originates from the swarm development~\cite{Kornienko_S05a} (see more in~\cite{Kernbach08}), many technologies and ideas from Jasmine platform is used here~\cite{Kornienko_S04b}. For the MAV development we utilized two additional sensors such as gyroscope and ultrasonic sensors for the flight stabilization~\cite{Suh03} and analyzing the surrounding environment. Therefore sensor reliability is imperative, because false readings could result in unavoidable crashes, leading to damages. In further developments the concept of collective embodiment will be used~\cite{Kornienko_S05e} as well as more complex sensors such as cameras~\cite{Mahony06}, however they require powerful on-board hardware to react quickly.

Depending on the task, the computational requirements are important for an autonomous flight since reaction times are essentially shorter than for ground vehicles. Additionally, the micro controller has to perform various time-critical tasks in on-board and on-line manner~\cite{Kernbach08online}, such as e.g. decision making~\cite{Kornienko_OS01} and a flight stabilization. Here, an application of a low-complex numerical approaches~\cite{Levi99} for decision making and planning~\cite{Kornienko_S03A}, \cite{Kornienko_S04} is required.

This work is organized in the following way. Firstly, in Sec.~\ref{sec:hardware} we shortly describe the developed hardware and software system. Sec.~\ref{sec:problems} is devoted to encountered problems, which required a special attention in further development. Finally, several performed experiments are demonstrated in Sec.~\ref{sec:experiments} and we conclude this work in Sec.~\ref{sec:conclusion}.

\section{Hardware and Software}
\label{sec:hardware}

\begin{figure}[ht]
   \centering
   \includegraphics[width=.49\textwidth]{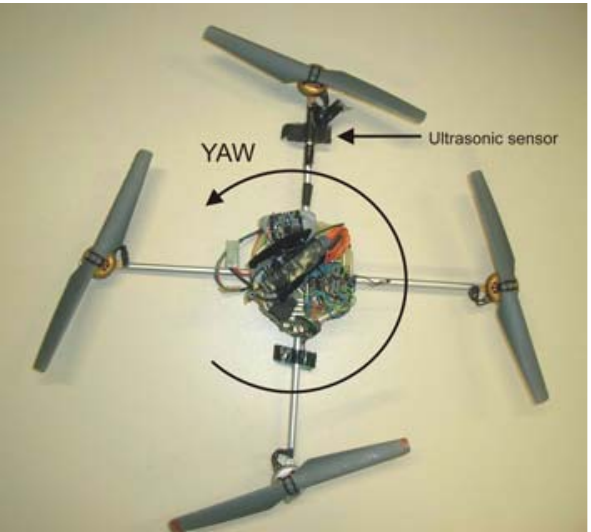}
   \caption{Prototype of the MAV. \label{fig:hardware}}
\end{figure}

The hardware design of the MAV~\cite{Jebens07} corresponds to the state of the art approaches for the quadrocopter-type of the aerial vehicles. There are four bars with four motors attached to the respective end of the bars. The other ends are held together in the middle, so that two bars are perpendicular to the others. All electronic components should be located in the center since a center imbalance will have only minimal side effects, see Fig.~\ref{fig:hardware}.
Components in the outer regions will block parts of the propeller surface lowering energy efficiency. The bars attaching the motors also have an underestimated effect on the flight properties: depending on the material it might suppress or increase vibrations induced by the motors. The weight of the bars will also have a much greater effect on the torque inertia than masses located in the center.

This quadrocopter has been designed using the following components:
\begin{itemize}
\item four 17cm aluminum pipes
\item four AXI 2254 brushless motors
\item two clock-/counterclock- wise propellers
\item four conrad brushless regulators
\item three ACT Micro Digital gyroscopes
\item one ADXL203 dual axis accelerometer
\item six SRF10 ultrasonic modules
\item two Jasmine mainboards with AT Mega168 micro controllers
\item one Graupner RC receiver
\item 3 Cell LiPo battery 1250 mAh
\end{itemize}

\begin{figure}[h!]
   \centering
   \includegraphics[width=.38\textwidth]{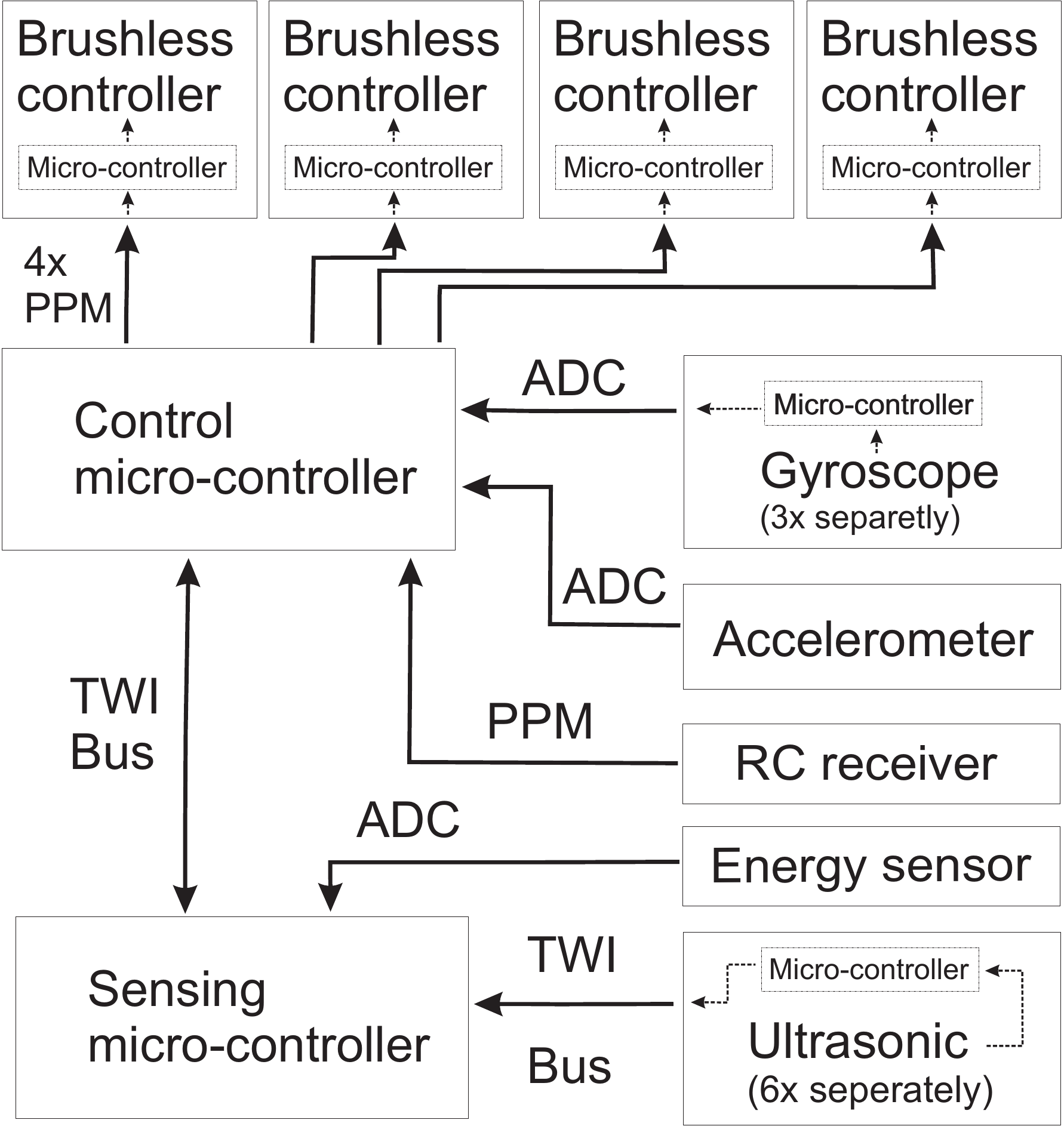}
   \caption{Structure of the electronics of the MAV.\label{fig:electronics}}
\end{figure}

Overall dimensions are 48 cm in diameter, maximum height in the center is about 10 cm. For the gyroscopes, the accelerometers and connections to the brushless regulators, ultrasonic modules, the RC receiver and two Jasmine mainboards holding the two micro controllers, a PCB was designed. The structure of the electronics is shown in Fig.~\ref{fig:electronics}.
It was decided to use mainboards from the micro robots Jasmine~\cite{KornienkoS05d} since they have proven to be a very stable platform in various experiments. Since we intend to study collective behavior of such MAVs~\cite{Kornienko_S04a}, the already existing software for these mainboards can be re-used.

As a result only the inputs and outputs had to be connected to these boards. The constructed mainboard provides a reliable platform for the core components and makes it possible to connect different sensor modules using TWI bus or switching to different receivers etc. The mainboard also has functionality to measure voltage level and current supplied by the battery. Especially the voltage measurement is highly advisable since a deep discharge will permanently damage lithium polymer cells. Since it is rather difficult to find clock and counter clockwise propellers, propellers from the model toy X-Ufo have been used. These propellers are relatively efficient and due to their soft material not as dangerous as most other model propellers.

The quadrocopter has been equipped with a ZigBee module for a wireless communication between a host computer. It allows observing flight parameters and their modification while being in the air. Since the ultrasonic modules are connected to the TWI any processor can directly control these modules. However, the second processor takes care of this task to reduce load on the control processor. The second processor performs calculations based on this information and just sends a suitable course correction command to the processor for flight control. So far calculations are limited to throttle commands dependent on the current height or roll commands if obstacles are in the way. For both calculations PID controllers are used.

\subsection{Providing stable flight}

\begin{figure}[ht]
   \centering
   \includegraphics[width=.49\textwidth]{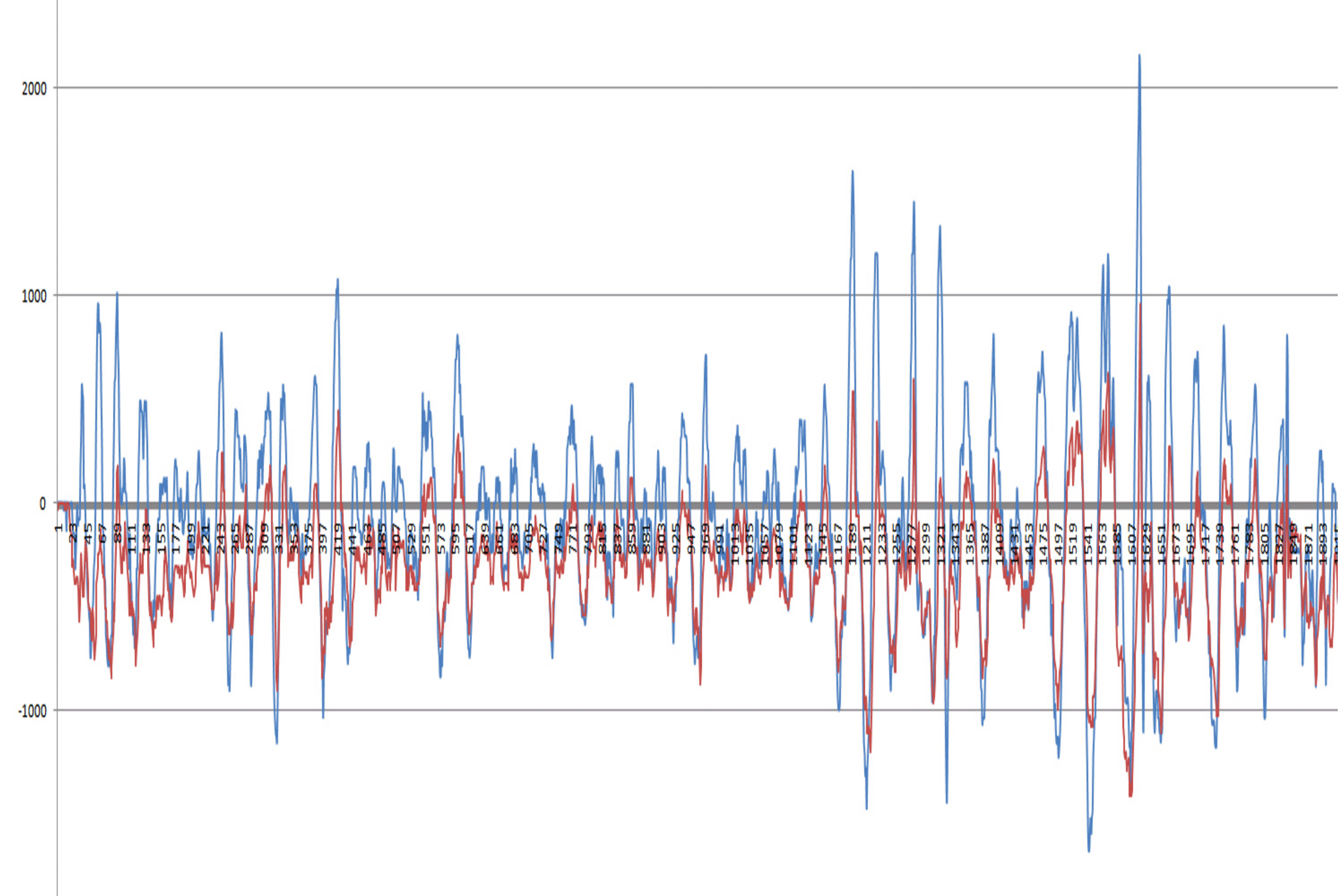}
   \caption{The roll gyroscope data during flight (red) and the corresponding course corrections (blue).}
   \label{img:gdata}
\end{figure}

On a macroscopic layer the first step can be seen as reading sensor signals, remote control commands and producing the suitable output for the brushless controllers. The output is calculated using three independent PID regulators for stabilizing the nick, roll and yaw axis. Assuming the model is symmetric. the nick and roll axis are stabilized using the same parameters. The Fig. ~\ref{img:gdata} shows the roll gyroscope data during flight (red) and the corresponding course corrections (blue).

Although this task is simple from a macroscopic point of view, the implementation on a micro controller is critical. Standard modeling components use pulse pause modulation (PPM) to provide signal information. In this case the RC receiver and gyroscopes will produce those signals and the brushless regulators are required to be controlled by PPM. To lower the complexity the gyroscope sensors were directly connected to the ADC ports of the micro controller. So there are four output PPM signals to be generated and four input PPM signals have to be read simultaneously. The input PPM signals from the RC receiver are the requested throttle, yaw, roll and nick commands. The input PPM signals are received 50 times per seconds resulting in 400 signal changes on the input ports per second. The output signals should be produced as quickly as possible since they are directly responsible for the regulation frequency of the entire system. The maximum frequency for the brushless regulators however is reached when the low period is reduced to zero. In practical this refers to a regulation frequency with model brushless regulators of about $500 Hz$. The difficulty is that 400 signal changes per second have to be measured with a precision of about $40 \mu s$ and simultaneously about 3000 output signals with the same precision have to be produced. Additionally, the calculations for the PID regulation have to be performed simultaneously. The main difficulty is keeping the precision during all thinkable events. Worst case events are for example a signal change on an input port triggering an interrupt and four timer events, which determine a signal change for the output signals. It is imperative, that any critical calculation during such an event is performed quickly, while non critical calculations are delayed. Otherwise there would be unacceptable false measurements or incorrect output signals.

\subsection{Improving flight stability}

To improve the hovering capability a dual axis accelerometer ADXL203 from analog devices has been used. While hovering at one specific spot the data from the accelerometer can be used to stabilize the quadrocopter in an upright position. However, since translational cannot be distinguished from gravitational accelerations, a sensor fusion of gyroscopes and accelerometers is imperative. Additionally accelerometers tend to be very sensitive to vibrations - much more than gyroscopes. It has therefore been decided to implement a kalman filter, which estimates the current orientation based on the gyroscopes and accelerometers. Using this information the translational accelerations can be computed as well. Additional PID regulators can be implemented to further stabilize the position.

\section{Problems}
\label{sec:problems}

\subsection{Timing}

In contrast to ground vehicles timings are very critical on board a quadrocopter, as several calculations have to be done in realtime. A delay would result in a crash. So for all actions the restrictions have to be carefully considered. Timings are especially critical since several time relevant actions happen simultaneously.

Among those are sensor signal changes which have to be evaluated between 100 and 1000 times per second. Steering commands using standard hardware require a signal analysis about 200 times per second. This is followed by the calculations, which produce the output signals for the four brushless regulators simultaneously.

Since those signal changes often have to be read and produced with considerable precision, other calculations have to be carefully moved to free time slots. The simultaneous actions not only require a powerful hardware, but also careful prioritization of tasks. The opacity of simultaneous actions and reactions makes the programming a serious challenge and requires exact planing.

\subsection{Sensor data}

Sensor signals on board a quadrocopter can be divided into important time critical and additional sensor signals. Time critical sensor signals can further be prioritized. The most critical data comes from gyroscope sensors since a loss of signal will immediately result in a crash of the quadrocopter. Less important are accelerometers which can be used to compensate quite disturbing drift effects of gyroscope sensors. A loss of accelerometers will result in a considerable, unnoticeable drift in position. At slow speeds ultrasonic sensors can be considered as a low priority sensor system which only has to be updated several times per second. Naturally the time frame decreases with increased velocities of the quadrocopter. However, sensor losses or considerable false readings have to be handled with extreme care independent of the sensor priority, otherwise resulting in false actions.

\subsection{Energy}

The energy requirements on board a flying platform are considerably higher compared to ground vehicles. With currently available high power lithium polymer batteries flight times of about 45 minutes can be reached. However, strict optimizations are required. For further increased flight times the design of the entire flight platform would have to be changed, however helicopter platforms are all of similar efficiency. In fact a quadrocopter is a rather energy efficient helicopter platform due to the low weight. Ordinary Sikorsky helicopters need a far more complex hardware design, which results in an exceptionally energy expensive increased weight. In general apart from the propulsion efficiency, all optimizations for overall energy consumption should almost solely be in the form of weight optimizations. An increased energy consumption of electronic components will normally be negligible in comparison to the increased propulsion energy needed to carry an additional gram. The high energy demand can be explained by the following formula:
\begin{equation}
F = \sqrt[3]{2*\rhoup*A*P^2}
\end{equation}
where $F$ refers to the thrust produced by an ideal propeller covering an area $A$ driven at a shaft power $P$ in a gas with density $\rhoup$ \cite{Schenk07}. In reality there are additional energy losses due to less than ideal propellers and energy efficiency will especially for small motors often be below 80\%. Probably the easiest factor for optimization, is the size of the propellers, if the overall size of the quadrocopter does not matter. The formula also describes that the thrust increases sublinear to the power input. Therefore any additional weight will become more expensive. Due to those weight restrictions all sensors not only have to be carefully chosen for their quality, but as an important factor also for their weight. If suitable miniaturized sensor and actuator modules are available, power demands decrease for small platforms, since $F \sim A^{\frac{1}{3}}$, $F_{required} = g * m$, where the mass $m \sim A^{\frac{3}{2}}$, however since energy reserves also decrease cubical, miniaturization is a very serious problem.

\begin{table}
\caption{Energy consumption} \label{table_example}
\begin{center}
\begin{tabular}{|c||c|c|}
\hline
Component & Energy consumption & Weight in \%\\
\hline
motors & 45W & 26\%\\
gyroscopes & 150mW & 4\%\\
accelerometers & 5mW & 0.1\%\\
ultrasonic sensors & 500mW & 5\% \\
Jasmine mainboards & 60mW & 5\% \\
\hline
\end{tabular}
\end{center}
\end{table}

\subsection{Hardware interaction}

As in most complex systems there are various side effects taking place. Those effects can be divided into EMC (Electro Magnetic Compatibility) problems and interactions due to mechanical coupling.

\subsubsection{EMC}

Various problems difficult to locate are related to EMC problems. Since the problem is increased with high currents the on-board hardware has to be carefully designed to protect sensor signals and micro controller hardware. Without a proper EMC planing it will be impossible to have a well functioning MAV. It is problematic, because the motors are driven by a considerable current, which is not constant in nature. As a result the induced currents can be quite high in all kind of wires on the PCB. Electromagnetic waves can also disturb micro controllers and sensor modules. Additionally any on board wireless communication might be interfered. Due to the high load on the battery pack there will also be a drop in voltage depending on the flight maneuver.

\subsubsection{Mechanical coupling}

The second difficulty arises due to mechanical effects. Since there are relatively strong forces and masses rotating at very high speeds the entire construction is vulnerable to strong vibrations. Those vibrations are responsible for serious false readings especially of accelerometers. It is therefore important to either lower the vibratory effects or to filter the sensor data accordingly. A well function IMU (Inertia Measuring Unit) requires a rather powerful and optimized filtering to gain reliable information.

\section{Experiments}
\label{sec:experiments}

Experiments have shown, that the software implementation is critical on a microcontroller and that software errors can produce harmful effects. The first reason is, that false measurements can for example result in a full throttle command although it has not been issued. Secondly, lost or delayed timer events will send a full throttle command to the brushless regulators. If such failures occur during flight time it will result in a serious crash or depending on the propellers can seriously harm human tissue. Therefore an eye protection has to be worn at all time. One has to consider that the maximum constant power of each motor is about 100W and that the peak power can be multiple times higher for a short period of time.

Flight experiments have shown, that it is not trivial to adjust the PID parameters even to lift off the ground. If the parameters are too much from the actual values it is impossible to hold the quadrocopter even approximately at the same position, i.e. it will wildly perform unacceptable course corrections. It has however been found out, that it is possible to find flyable parameters using solely a PD regulator. However, adjusting the integral part considerably improves flight performance. Unfortunately those parameters cannot be optimized one by one since they influence each other. Naturally the regulation frequency is also an important factor. Experiments have shown, that increasing the frequency improved flight characteristics considerably up to $300 Hz$. At such regulation frequencies the quadrocopter hovers extremely calm.

Nevertheless, even a perfect trimming to the hovering position results in a relatively quick translational movement. Therefore the pilot has to perform constant course corrections. The situation makes it difficult to implement an autonomous navigation. Further experiments have shown, that a PID regulation based on the estimated orientation using the data calculated from the kalman filter greatly improves flight stability. Holding a certain position becomes much easier, since the drift is reduced significantly. Although such a sensor combination is not able to absolutely hold a set position drift rates become so slow that a low speed sensor system can regularly move the MAV to its original position. This makes it much easier to navigate autonomously with for example ultrasonic sensors.

\begin{figure}[ht]
   \centering
   \includegraphics[width=.49\textwidth]{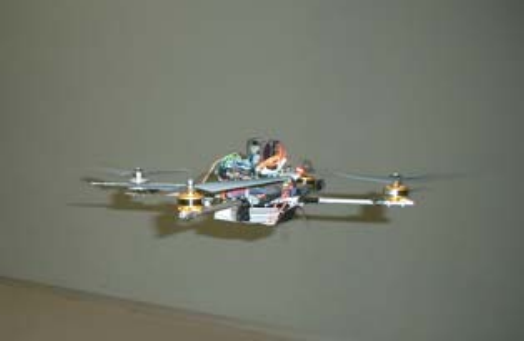}
   \caption{Quadrocopter during a flight.\label{img:flightheight}}
\end{figure}

During the experiments the ZigBee module proved to provide an extremely robust communication between a ground computer and the quadrocopter over an indoor distance of about $30 m$. The data has been transmitted at a speed of $19200 baud$. Changing regulation parameters during flight proved to be very helpful. Finding stable parameters has been very time consuming and can now be accomplished in a matter of minutes. However, in several situations it is prudent to test a change in the regulation system, e.g. a limitation factor or a sudden regulation jump. Another advantage is, that that sensor data can be directly transmitted to a ground station for analysis. Otherwise, the data would have to be saved on board and later transmitted. Since the RAM is limited in size it does not serve as a solution. All in all, experiments have shown, that it is strongly advisable to use a wireless on-board solution.

\subsection{Flight at a Constant Height}

\begin{figure}[h!]
   \centering
   \subfigure[\label{fig:height1}]{\includegraphics[width=.24\textwidth]{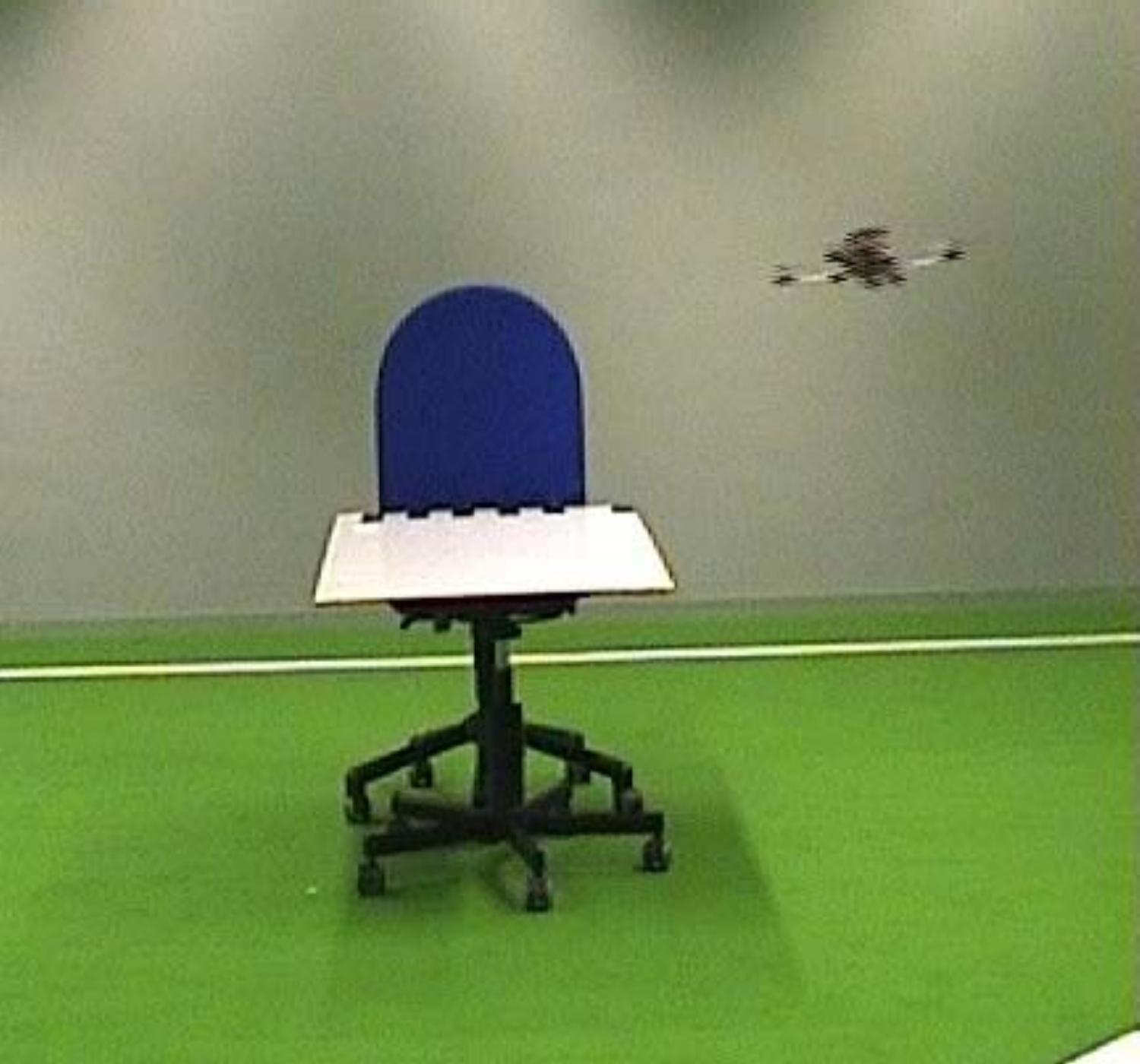}}~
   \subfigure[]{\includegraphics[width=.24\textwidth]{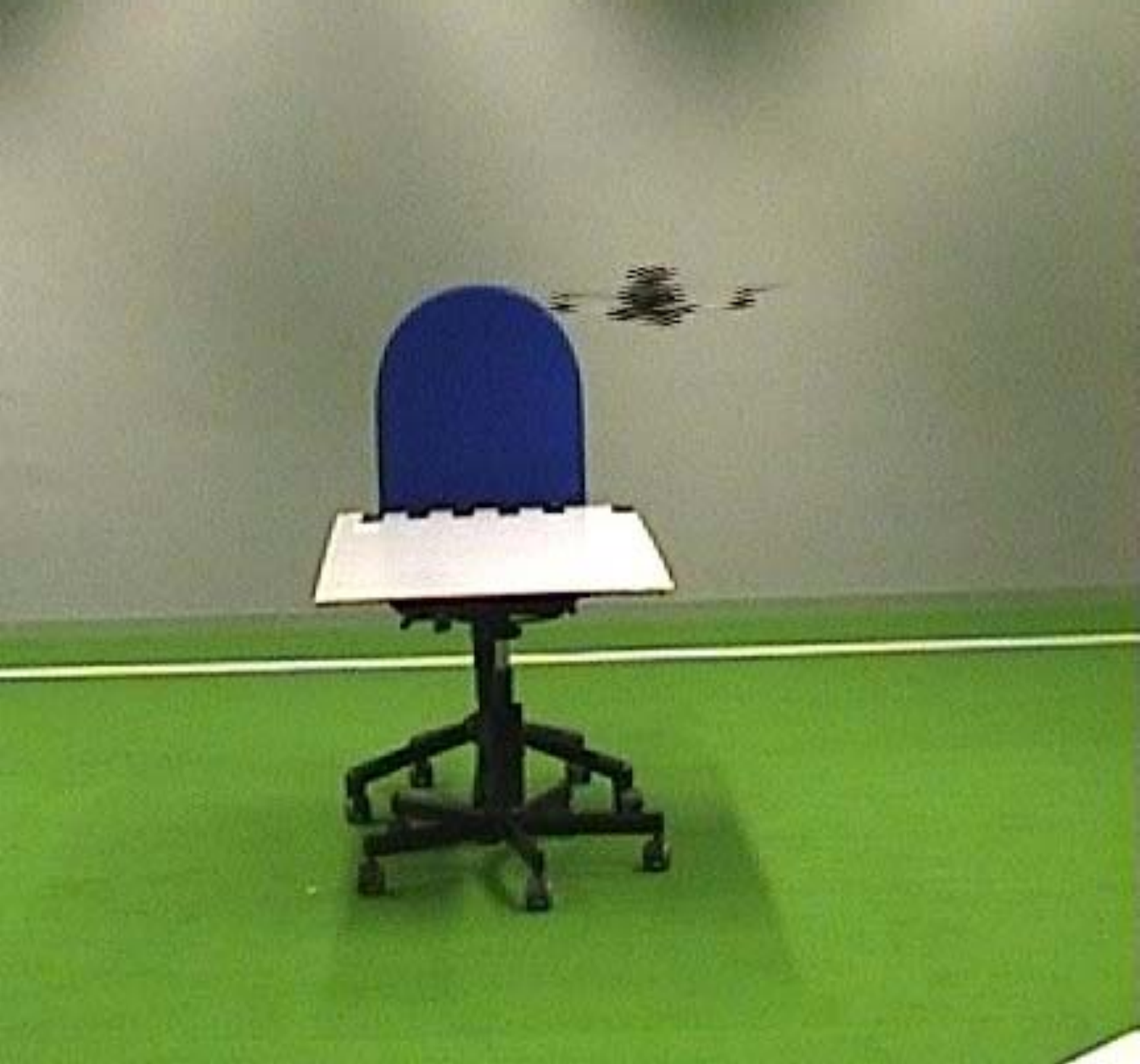}}\\
   \subfigure[]{\includegraphics[width=.205\textwidth]{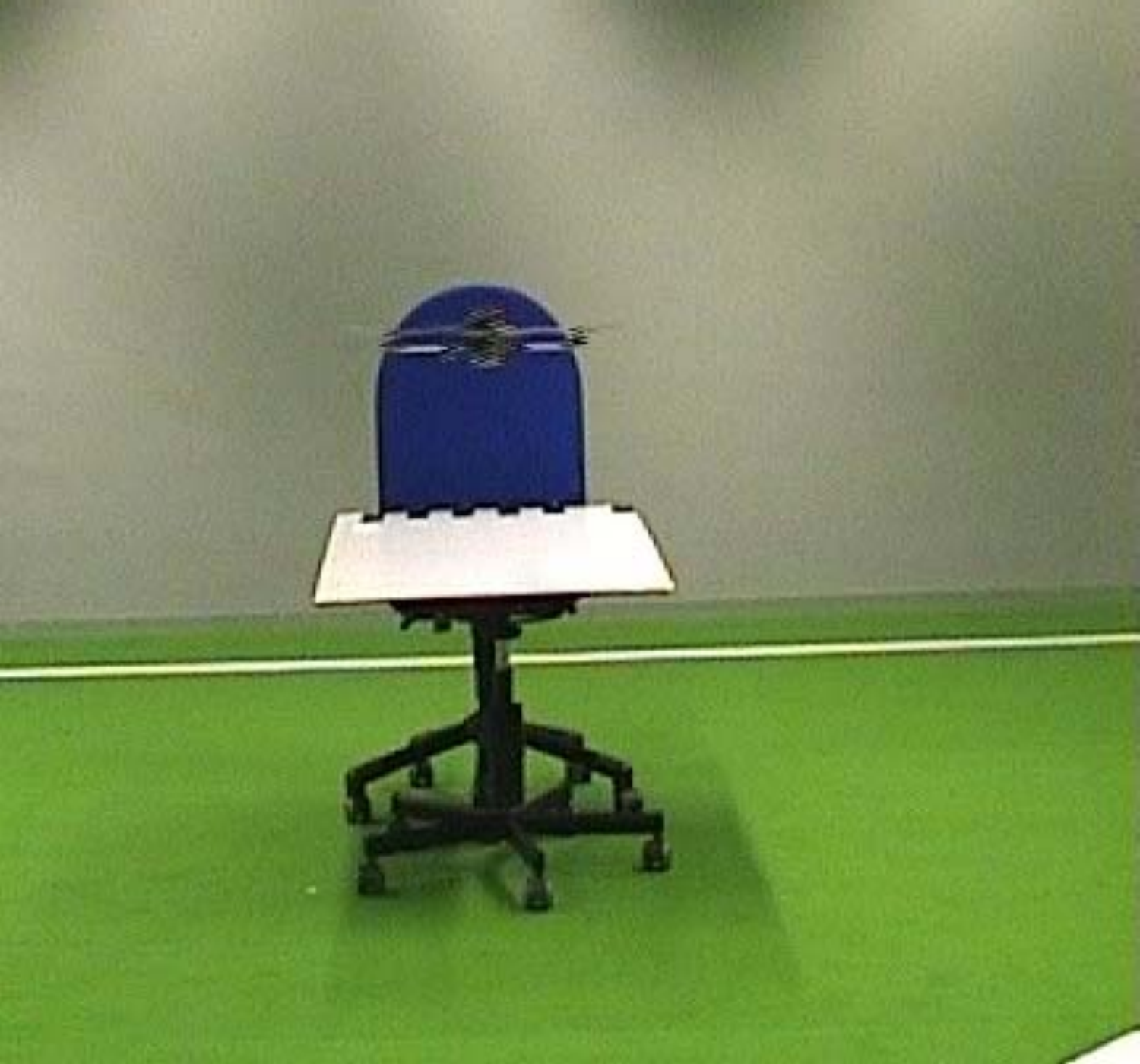}}~
   \subfigure[]{\includegraphics[width=.24\textwidth]{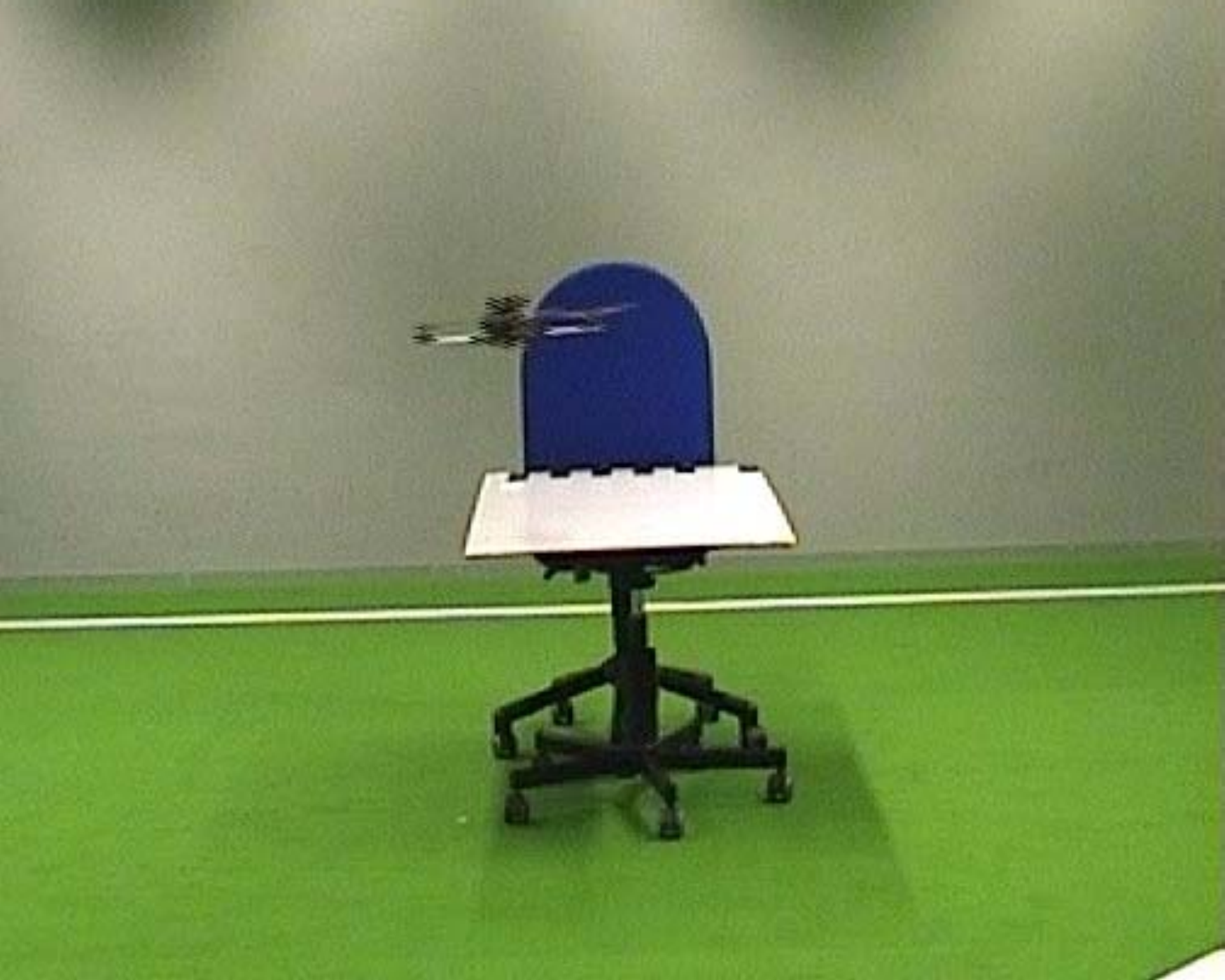}}\\
   \subfigure[]{\includegraphics[width=.24\textwidth]{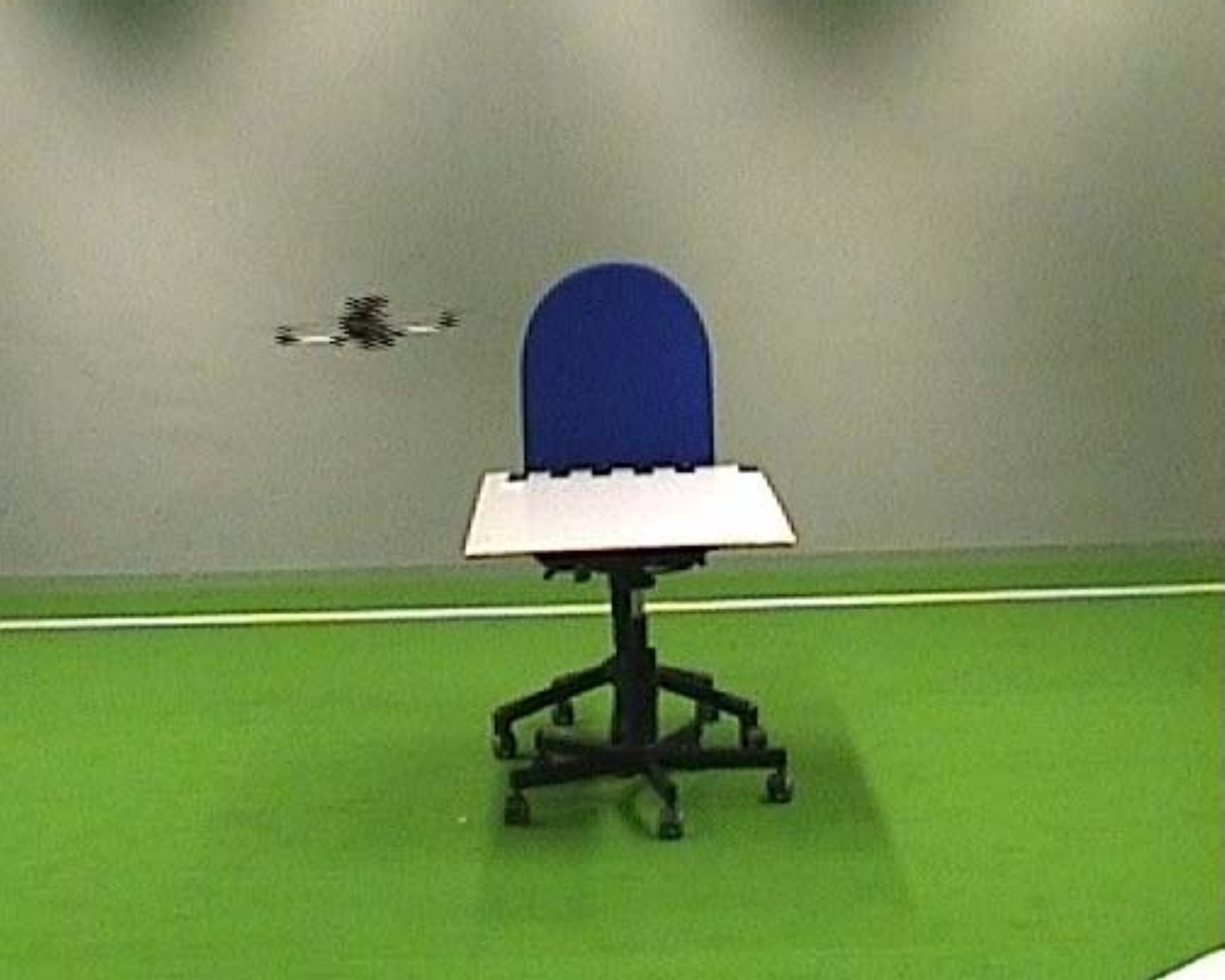}}~
   \subfigure[\label{fig:height6}]{\includegraphics[width=.24\textwidth]{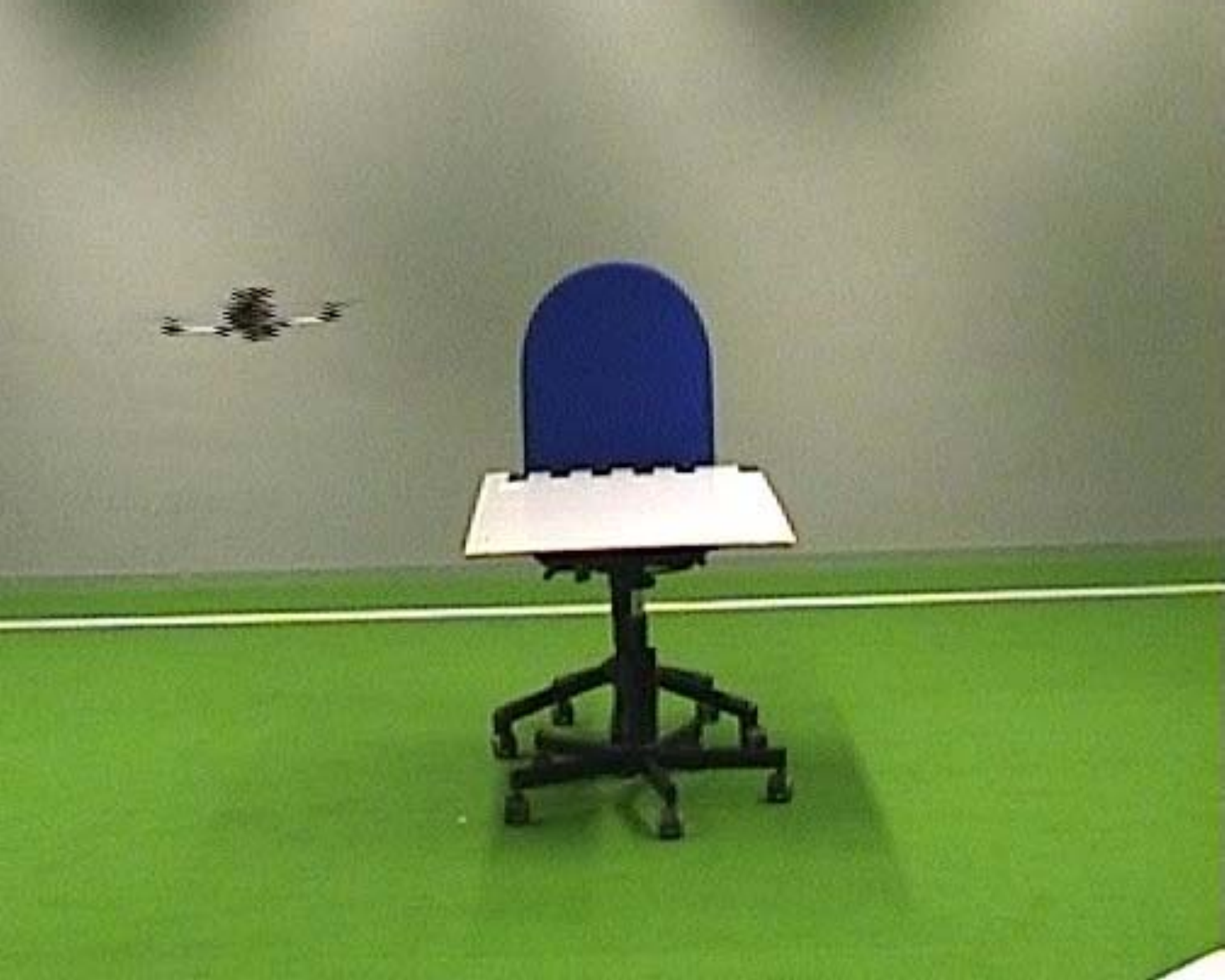}}\\
   \subfigure[\label{fig:height7}]{\includegraphics[width=.49\textwidth]{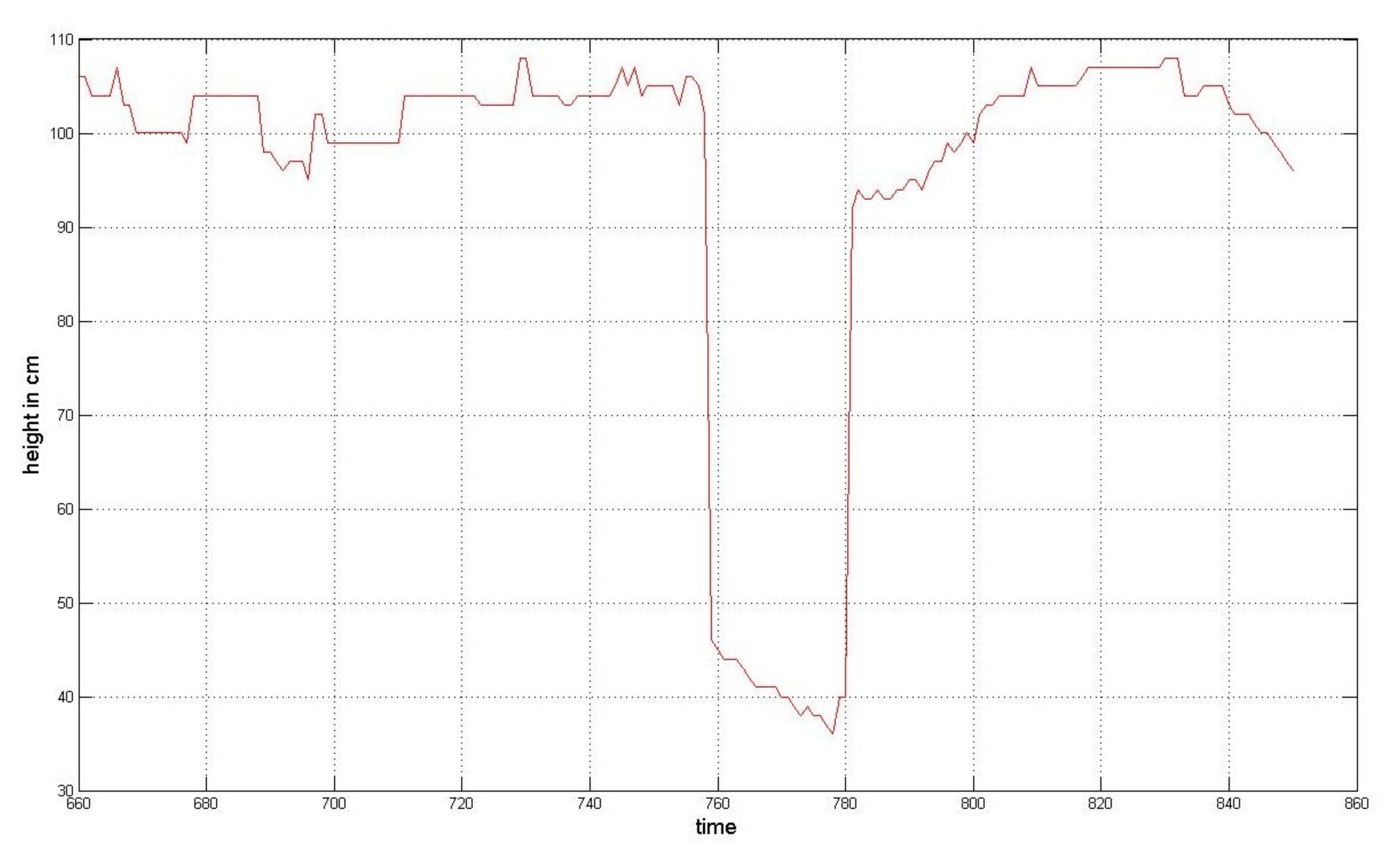}}
   \caption{\textbf{(a)-(f)} Experiment with a flight at a constant height; \textbf{(g)} The measured height level. All images are from \cite{Jebens07}.\label{fig:height}}
\end{figure}

The experiment of a flight at a constant height, see Fig.~\ref{img:flightheight}, intends to demonstrate capabilities of sensor-data processing from  the gyroscopes, accelerometers, ultrasonic sensors and the reactions sent to the brushless controllers. Figs.~\ref{fig:height1}-\ref{fig:height6} show a flight in a relatively constant height referring to the ground over a chair.
Fig.~\ref{fig:height7} shows the sensor data of the ultrasonic sensor captured during this flight. The flight over the chair can clearly be seen as a height drop of about $60 ~cm$. The overall data is relatively smooth, the distance fluctuations can also be seen in the experimental data.

\subsection{Collision avoidance}

\begin{figure}[ht]
   \centering
   \includegraphics[width=.5\textwidth]{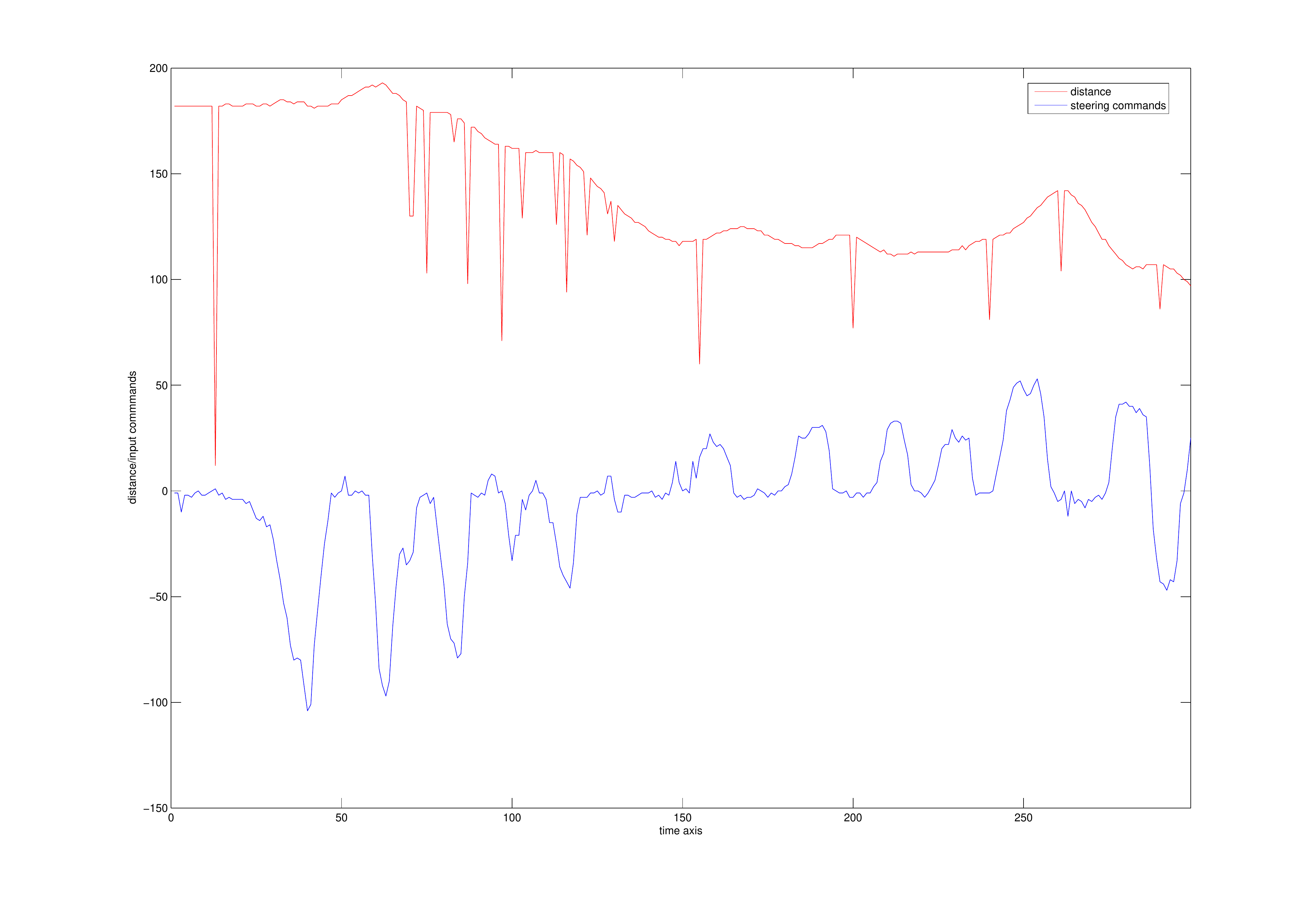}
   \caption{Distance to a obstacle (nearby wall). The green line indicates the distance to the wall, the yellow the steering commands of the pilot}
   \label{img:obst}
\end{figure}

Several additional ultrasonic sensor modules enable the quadrocopter to not only hold the height, but also to implement a 3D collision avoidance. For the first experiment only the roll axis was regulated by a sensor signal from one side. The PD regulation was able to keep the quadrocopter within a certain distance to the wall and by adding a sort of emergency factor it is very unlikely for a pilot to hit objects. However, it was not possible to find parameters holding a quadrocopter in any premature situation constant at a certain distance. When the aircraft approaches the wall too quickly, the used parameters were not good enough to keep it away from the wall unless there is a strong emergency factor involved. However, such an abrupt manoeuver introduces a rather chaotic flight behavior the next couple of seconds, which can only be secured by a human pilot. Fig.~\ref{img:obst} demonstrates a human pilot roughly holds the distance to a nearby wall. The green line indicates the distance to the wall, the yellow the steering commands of the pilot.

Using six ultrasonic sensor modules pointing in all directions it should be possible to implement a 3D collision avoidance capable of navigating safely in an unstructured indoor environment. This kind of sensor system can be supported by a camera system, to further increase flight safety and to perform higher tasks. However, it has
to be mentioned, that implementing a stable 3D collision avoidance requires careful parameter tuning, otherwise the quadrocopter will perform flight maneuvers, which cannot be re-stabilized in time by ultrasonic sensors due to the relatively low update speed.

\subsection{Autonomous Flight}

\begin{figure}[ht]
   \centering
  \includegraphics[width=.5\textwidth]{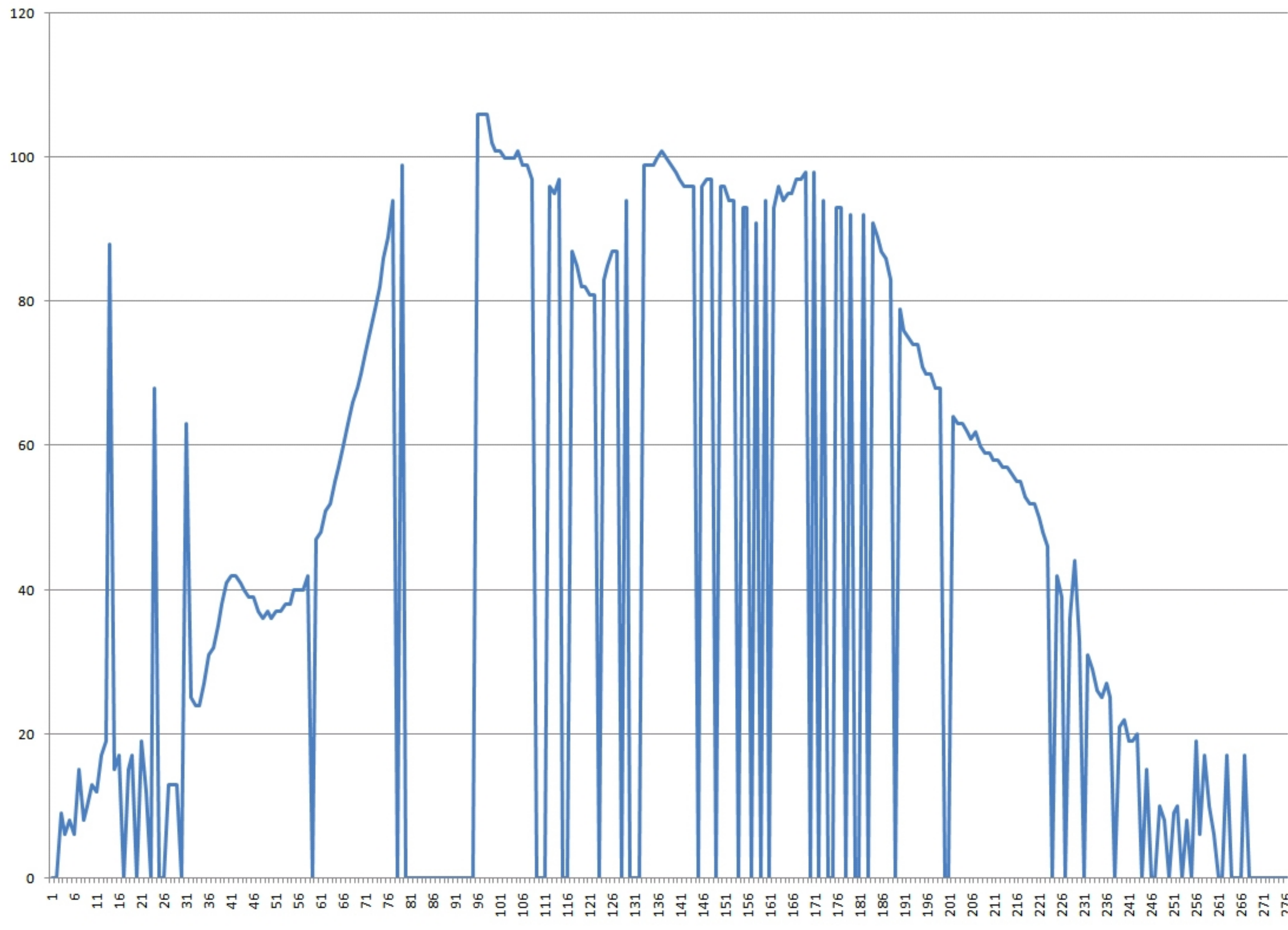}
   \caption{Height profile of start and landing.}
   \label{img:height}
\end{figure}

Since flight performance has been tuned to a very stable level, the next task was implementing an automatic flight control with the use of an ultrasonic module. The goal was a height control, permanently adjusting the throttle to stay at a given height level, and collision avoidance. Even in this experiment it turned out that finding suitable parameters is not trivial. Dramatic problems can occur due to faulty measurements. For example, the regulation needs to be turned off as soon as there is no signal received from the sensor module or the throttle will attain to a faulty value. Fig.~\ref{img:height} visualizes a height profile of start and landing. It can be seen that the values are quite often interrupted by zeros, which refer to invisibility errors. It is important to treat those values with care. Due to the tilt of the MAV during flight time, there occurs another problem of measuring the height by ultrasonic sensors. This produces a height measurement, which is diverted from the real value and is therefore dependent on the tilting value and also on the opening angle of the ultrasonic scanning beam.

It was possible to implement a reliable height control with a simple PD regulator. This is deactivated upon zero measurements and has a limited throttle adjustment, therefore requiring the pilot to roughly adjust the throttle. Without the D factor the quadrocopter will swing around the requested height.

\section{Conclusion}
\label{sec:conclusion}

This work has been done within several exploratory pilot projects for extension of ground swarms into underwater~\cite{jellyfish}, reconfigurable~\cite{Kernbach08_2} and aerial~\cite{Jebens07} dimensions with the goal to explore adaptive, self-organizing and evolutionary approaches in different embodiments. Several of these pilot projects results later in the research projects~\cite{Kernbach08Permis}, \cite{Levi10}, \cite{cocoro}.

We successfully implemented the hard- and software of a cost-effective MAV capable of a 3D obstacle detection and automatic course corrections. This work proved that the used sensors are enough to provide a random flight within a partially unstructured environment. For further miniaturization the energy is the most prohibiting factor. Current sensor modules based on MEMS technology enable a quadrocopter size of less than 10 cm in diameter. However, without dramatically reducing the weight it will be impossible to keep such helicopters in the air for long time periods. Future prototypes need to have an improved hardware design to counteract various software issues. Brushless controllers need to be controlled via a common bus system or ideally they all should be integrated on one board. Furthermore, a sensor system with an integrated bus system could lower noise effects and rise overall flight performance. This could be achieved by using more enhanced micro controllers on board, which would also allow the use of additional software filters. Such a micro processor will also decrease software complexity for example by introducing the capability of threading. Especially with growing overall complexity, i.e. vision controlled flight, such a system is essential.

Experiments have shown, that computational requirements are quite enormous on board a flying robot. Although most calculations are mathematically not too complex, they have to be performed hundreds of times per second. Especially when more complex regulation models are used, much more powerful micro controllers are imperative. Future prototypes are planned to be equipped with a micro controller at a speed beyond 100 MHz and capable of handling several interrupts in parallel. Such a processor will enable the quadrocopter for example to track other object based on visual information.


\end{document}